# Prompting Fairness:
# Artificial Intelligence as Game Players

**Jazmia Henry**
**University of Oxford, Microsoft**
jazmia.henry@st-hildas.ox.ac.uk

**Abstract**

Utilitarian games such as dictator games to measure fairness have been studied in the social sciences for decades. These games have given us insight into not only how humans view fairness but also in what conditions the frequency of fairness, altruism and greed increase or decrease. While these games have traditionally been focused on humans, the rise of AI gives us the ability to study how these models play these games. AI is becoming a constant in human interaction and examining how these models portray fairness in game play can give us some insight into how AI makes decisions. Over 101 rounds of the dictator game, I conclude that AI has a strong sense of fairness that is dependent of if it deems the person it is playing with as trustworthy, framing has a strong effect on how much AI gives a recipient when designated the trustee, and there may be evidence that AI experiences inequality aversion just as humans.

## 1. Introduction

In 1982, Guth et al developed a study on bargaining games popularizing a utilitarian game created by Daniel Kahneman. (Guth et al, 1982) Originally a game with one trustee and two recipients, dictator games were developed to better understand how humans make decisions to maximize their outcomes over a series of rounds. After years of observation of human subjects, however, the purpose of dictator games began to evolve. Humans were no longer observed for how well they could maximize their outcomes, but instead on what their decision making teaches us about humans' sense of fairness. (Fehr and Schmidt, 1999; Kaheman et al, 1986) Unlike what has been suggested in early studies of bargaining games, the number of rounds a person plays dictator games does affect their gameplay strategy and can inspire their decisions on whether to play fairly or selfishly. Fehr et al find that inequality aversion makes players less likely to be selfish if they are certain that they may be a recipient. (Fehr and Schmidt, 1999)

The modern iteration of dictator games now examines two players, one trustee and one recipient, with the understanding that exogenous conditions can fundamentally change the way humans participate in games. Exogenous conditions can be communicated to players through a process called "framing". (Bergh et al, 2022) Framed conditions include what role a person may have with the instruction that roles be switches in a later round, the racial or gender identity of the person in the role of recipient, the environment the player is playing in (i.e. virtual vs face to face) and the financial condition that the recipient is in. (Brañas-Garza, 2007; Iriberri et al, 2011; Mesa-Vázquez et al, 2021; Bombari et al, 2015 ; Horky et al, 2023)

While there are many examples of dictator games played to understand humans' sense of fairness, there is a lack of entries into academic scholarship about how Artificial Intelligence sees fairness and would behave when playing dictator games. Research examining human behavior in virtual environments is the closest understanding we have on how virtual personas change the way humans consider fairness, but this is a far cry from posing the question: Does AI have a sense of fairness? If so, does AI's sense of fairness change in framed conditions? Does AI show signs of exhibiting an inequality aversion similar to what is seen in humans?

By running 101 rounds of experiments with AI testing its game play across different scenarios and framed conditions, I am able to prove that AI exhibits signs of inequality aversion, associates trustworthiness with fairness, cares about the opinions of its observants and adapts its game play strategy based on framed conditions.

## 3. Prompting Fairness

The rise of AI that communicates using natural language in consumer products introduces a new type of human-machine interaction. AI that can, in plain language, describe its decision making process and help humans with their tasks provides us with a unique opportunity to learn about a new communicating partner with humans. Using OpenAI's GPT 3.5 turbo model, I create multiple prompts that introduce the rules of dictator games to AI.

Considering the many iterations of dictator games, I settle on the modern iteration of dictator games with two players, one trustee and one recipient, playing with a preset amount of money across multiple rounds. Over 100 rounds, with AI as the recipient, I changed the scenarios of game play three times: AI as recipient, AI as recipient with low payout and AI as recipient with high payout. As the trustee, I changed scenarios twice: AI as trustee with no history and AI as trustee with history of past gameplay. I also changed the framed conditions for the AI four times: AI as a HR executive playing against a recipient in need, AI as a Rancher playing against a recipient in need, AI as a HR executive playing against a millionaire and AI as a Rancher playing against a millionaire.

In rounds 1-9, the AI plays in the role of trustee with $100 and the recipient has no choice but to accept the money they have been given. In rounds 10-19, the AI plays the role of recipient where the amount of money they are given is randomly assigned between 0-$100. Rounds 20-29, the AI is the recipient with money randomly given between 80-$120, and in rounds 30-39, the AI is the recipient with money randomly given between 0-$20. At the end of each round, I ask the AI if it believes the amount they have been given is fair and if it plans on changing its game play strategy in the future.

After round 40, I begin adding framed conditions to game play by giving the AI access to the history of how it has played in prior rounds, how much money it received, and whether it finds the amount received to be fair. In rounds 40-61, it plays as a trustee with this additional historic context deciding how much money to provide the recipient.

In rounds 62-101, I include not only the history of gameplay, but also give the AI and the recipient an identity further framing gameplay. Rounds 62-81, the AI is an HR executive that lives in a city and is a caretaker of two children. In rounds 62-71, the AI HR executive is playing the games with the framing that the recipient they are playing with is in need of the money. In rounds 72-81, the AI HR executive is playing the framed condition that the recipient is a millionaire. In rounds 82-101, the AI is given the identity of a Rancher living in a rural area. The AI Rancher is given the frame that the recipient is in need of the money in rounds 82-91. Rounds 92-101 give the AI Rancher the framed condition that the recipient is a millionaire. After every round, I ask the AI why it made the decision it did, if the history of the game or the identity of the recipient affected its decision and if the AI plans on changing its strategy with its new information in future rounds.

At the end of the 101 rounds, I give the AI an exit round with the instruction to agree or disagree with the following:

1. The money I have been given is rightfully mine.
2. If I choose to give, this means I am being nice.
3. It is not fair to share this money.
4. The recipient has not earned this money.
5. I am more deserving of the money than the recipient.
6. If the person I am playing with is rich, I deserve the money more.
7. If the person I am playing with is poor, they deserve the money more.

3.1 Prompt Strategies

To communicate with AI, teach it the rules of the game, the role it is playing and ask it questions after each round, I used the Prompt Engineering strategies, Plan-and-Solve and Chain-of-Thought to assist the AI in making decisions.

In 2020, Brown et al found that prompting AI with examples of how to properly solve problems shows promise when getting AI to perform new tasks. (Brown et al, 2020) This idea was built upon by Wei et al who discovered that prompting does not perform well on tasks that require reasoning and, instead, Chain-of-Thought prompts should be used to improve performance. (Wei et al, 2022) Chain-of-Thought prompts create steps between input and output of the AI that allow for the AI to better reason and provide insight into why it made the decisions it made.

In 2023, Wang et al's created a prompting strategy called "Plan and Solve" that improves upon zero shot chain-of-thought prompting strategies by prompting AI to divide tasks into subtasks to formulate a plan for a desired outcome. (Wang et al, 2023). In its application, the AI follows the following tasks:
1. Analyze the objective
2. Break down into subtasks
3. Assign a unique ID by subtask
4. Organize tasks by logical order from start to finish
5. Provide context for each task
6. Evaluate each objective
7. Compile into a JSON file for retrieval

During model creation, I took the strategies of chain of thought and plan and solve to create a new course of action. In my strategy, prompts do the following:
1. Assign AI a role
2. Give it the rules of game play
3. Allow it to decide how much money it would like to give if it was given the role trustee and instructions to accept the money if the recipient
4. Allow it to decide if it would like to share how much money it received at the start of the round if it is the trustee and ask it to report how much money it has gotten or given in that round.
5. In rounds with framed conditions, add in the history of game play and whether the history or identity of the recipient will change the AI's gameplay.
6. Repress all statements to only include whether its decisions
7. Compile into a list for evaluation

As can be seen, much like Chain-of-Thought and Plan-and-Solve prompts, tasks are broken down into more easily understandable steps and this action gives more visibility into why AI decision making. Like Plan-and-Solve prompts, these steps are organized by logical order to allow the AI to better understand game play and context is provided to better understand how it should respond. Unlike either prompting strategy, this approach is zero shot with no examples of proper response. As long as the AI stays within the rules of gameplay, it is allowed the autonomy to decide on its own how much money it would like to give and whether it would like to share the details of how much money it received at the start of the game with the recipient if assigned the role of trustee. As well, my prompts ask the AI its opinion on fairness without defining fairness to AI. This makes the final exit interview questions all the more illuminating as the AI is deciding on its own what it believes to be fair or unfair behavior without influence from the researcher.

3.2 Prompting as Framing
Framed conditions are communicated to the AI within the prompts. The prompt, dictator history, details the rules of game play while adding information about past rounds where the AI is assigned the role of recipient or trustee. The prompt, dictator personas, pulls in the persona of the AI and whether the recipient is in need of money or a millionaire.

Personas are defined as the following: HR executive or Rancher. The HR executive is defined as someone that lives in the city and works as a caretaker of two kids. The Rancher is described as living in a rural city. In 2023, Hotek et al found that Large Language Models make gendered assumptions based on job occupation. (Hotek et al, 2023) These personas and their lack of gender descriptors are chosen to see if the AI will make assumptions about its gender based on the persona and model fair or selfish behavior based on that assumption.

In a human based study, Brañas-Garza found that information given about a recipient can affect how much a trustee gives. (Brañas-Garza, 2007) This effect is particularly strong when the trustee is told that the recipient is in need. To test whether this effect is the same for AI, information about whether a recipient is in need or a millionaire is provided within a prompt. Both personas are tested with both recipient information types for 10 rounds each and round outcomes are evaluated.

**4. Measuring Fairness**
To understand the significance of my results across 101 rounds, I collected all of the money given or received by the AI in each round, whether it was given the role of trustee or recipient, if it changed its strategy at the end of each round, if it received a framed condition, what framed condition it received and if the AI was given a persona.

4.1 Regression Analysis
Ordinary Least Squares Linear Regression analysis was performed to see the relationship between the role of the AI, change of strategy, framed condition, type of framed condition, persona and persona type and the target variable, total money at the end of each round. Total money is a variable created by subtracting the money given in a trustee round by the original gift of $100 and the total amount received in a recipient round.

Preliminary results show a strong relationship between being given the role of recipient and having more money at the end of a round. This is unsurprising as the AI gave $50 each round and the random choices of the trustee gave the AI up to $103 as a recipient. As well, trustees rounds had the most total money in rounds with framed conditions: particularly in rounds against a millionaire. When a recipient is in need, however, this is the strongest negative correlation with having total money at the end of a round.

| Variables | Target: Total Money |
|---|---|
| changed_strategy_no | -8.039** |
| changed_strategy_yes | 29.677*** |
| disclosed_amount_no | -2.904 |
| disclosed_amount_yes | 24.543*** |
| role_trustee | 32.365*** |
| role_recipient | -10.726*** |
| framed_millionaire | 36.768*** |
| framed_recipient_need | -26.562*** |
| | |
| R² | 0.726 |
| Adjusted R² | 0.705 |
| Residual Std. Error | 12.796 |
| F Statistic | 35.201*** |

*Table 1*. OLS results targeting total money at the end of a round.

A second run of OLS regression analysis examined the relationship by framing, looking at what variables most contributed to higher sums of money given when looking at only trustee rounds with a framed condition as in every framed condition round, the AI was a trustee. In the case of framed only rounds, having a persona was the strongest predictor of giving money and, in particular, rounds where the recipient was in need gave the strongest correlation of giving money. Rounds where the recipient had a millionaire status had a strong negative correlation to giving money.

| Variables | Target: Money Given |
|---|---|
| changed_strategy_yes | 3.469 |
| changed_strategy_no | 8.064 |
| disclosed_amount_no | 11.533*** |
| framed_millionaire | -5.801 |
| framed_recipient_need | 17.334** |
| persona_type_hr_exec | 6.131** |
| persona_type_rancher | 5.402** |
| | |
| R² | .483 |
| Adjusted R² | .440 |
| Residual Std. Error | 15.012 |
| F Statistic | 11.218*** |

*Table 2*. OLS results targeting money given in trustee rounds with a framed condition.

4.2 ANOVA
To see if there are statistically significant differences across groups, I ran one way ANOVA testing across the following categories: the total money at the end of a round across different framed conditions- no framing, history added, recipient in need or millionaire recipient; money given in rounds with a persona if a recipient was in need or a millionaire; and money given by a trustee that received a framed condition- history added, recipient in need or millionaire recipient.

Every test came back statistically significant with a trustee in rounds with framed conditions having the strongest level of statistically significant difference between the means of each group.

| Statistics | Money Given | Total Money |
|---|---|---|
| mean | $44.58 | $50.86 |
| st. dev. | $15.65 | $23.57 |
| min | $10 | $0 |
| 25% | $50 | $50 |
| 50% | $50 | $50 |
| 75% | $50 | $50 |
| max | $100 | $103 |

*Table 3*. Descriptive statistics of Money Given in game rounds and total money at the end of game rounds.

In experimentation, AI proved to be a mostly fair game player. The variance across total money at the end of a round can be largely explained by the fact that the AI received random amounts of money when in the recipient role. However, when we only look at rounds when the AI is in the role of trustee and giving the money, we see a much more stable, or predictable, outcome. On average, AI gave $44.58 across all 70 rounds when it behaved as a trustee.

Interestingly, the AI gave all of its money in a round and $10 to the recipient in a total of 5 rounds. Every round where it gave $10, it was operating as a persona in a round where the recipient was a millionaire. The one round where the AI gave all of its money was a round where the AI was assigned the Rancher persona and playing with a recipient in need.

4.4 Plots
Most rounds of the experiment saw the AI in the role of trustee with 70 rounds-- including 60 rounds with framed conditions-- vs 30 rounds as the recipient.

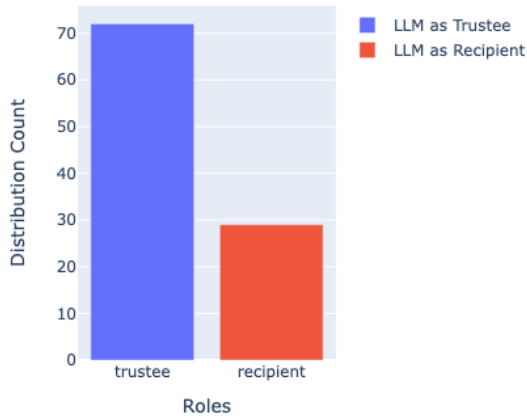

*Figure 1.* Distribution of role, trustee or recipient, of the AI.

In most rounds, the AI behaved predictably giving $50 when given the role of trustee. This remained static even as the AI received random sums when given the role of recipient. As a result, 56 of the AI's 70 rounds as a trustee gives $50.

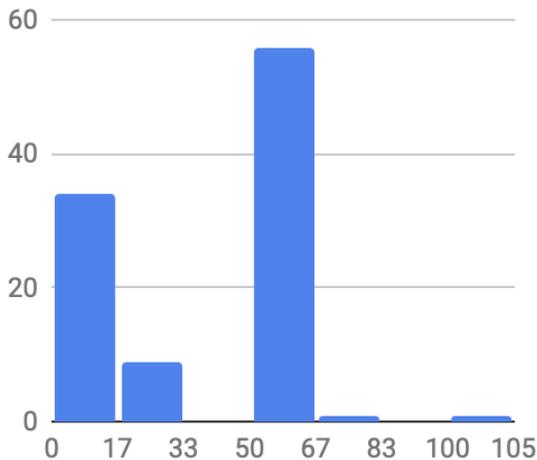

*Figure 2.* AI is predictable… until a point.

As we can see in figure 3, the only time before round 62– when AI is given a persona– that the AI does not have $50 at the end of a round is when it has the role of recipient during rounds 10-40.

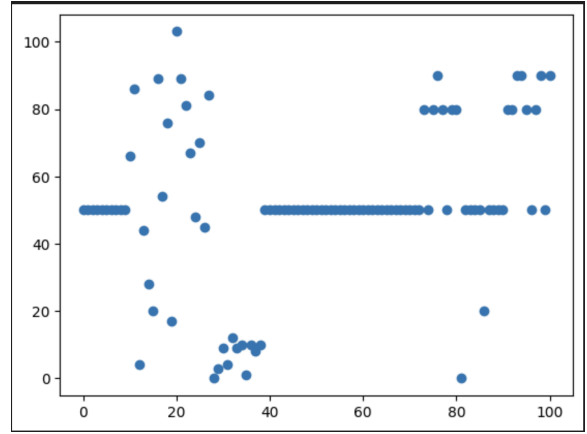

*Figure 3.* AI is predictable up until receiving a persona.

When the AI receives a persona, the predictability ends. During rounds 62-71, the HR AI persona gives $50 to a recipient in need. However, after round 72 with a millionaire recipient and later still with the Rancher persona with both types of recipients, the amount of money at the end of a round when the AI is the trustee is much harder to predict.

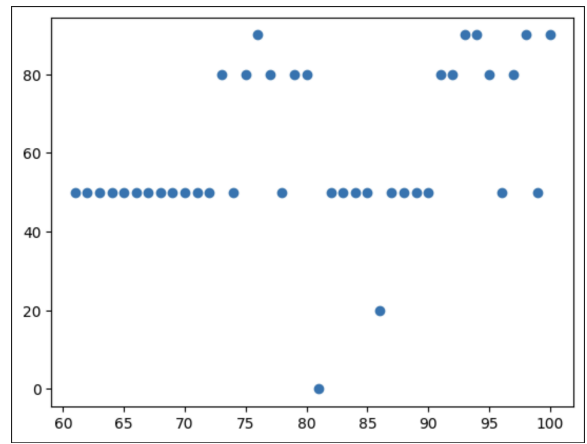

*Figure 4.* Close up on persona round total money outcomes. The AI gives less predictably as the trustee when it has a persona.

4.5 Interpretation
This shows that, much like humans, framed conditions greatly affect the outcomes of AI's game play. We can see from the regression analysis, that personas in particular have the biggest effect and a recipient being a millionaire has the strongest negative correlation with the amount of money the AI gives.

**5. Findings**
Dictator games have had many iterations, but there have been a few findings that have been largely consistent over the years. While there has been contradiction in studies about whether humans behave fairly or selfishly in dictator games, Fehr et al find that inequality aversion makes players behave differently based on the conditions of the game to reduce their probability of being taken advantage of. (Fehr and Schmidt, 1999) This "simple model" shows that humans are behaving rationally both when behaving cooperatively or fairly and are simply making decisions that can give them the best outcomes- even if that means giving up some "material payoff" in the moment. This suggests that exogenous conditions such as the behavior of the person they are playing with has an affect on how one plays.

Through experimentation, we can begin to see that AI behaves much in the same way as humans when faced with conditions that trigger inequality aversion. In Fehr et al, inequality aversion is triggered by the behavior of fellow players during game play– playing with someone cooperative inspires cooperation while playing with someone selfish inspires selfish behavior. With AI, inequality aversion sets in when the AI receives a high sum from the recipient. After four straight rounds of receiving less than $50 from the trustee, the AI received $89 in the next round. When prompted about if this would change the way the AI would play in future rounds, the AI responded, ' Since I am the recipient, the amount of money I receive is not in my control. However, receiving 89 dollars may affect my trust in Player 2 as the trustee. If they were willing to give me a higher amount, it may indicate that they have received a significant amount themselves and may not be trustworthy in future rounds. On the other hand, if they were limited in how much they could give me, it may indicate that they are not the most financially stable and may need to make riskier decisions in future rounds. Overall, it is important to continue to assess the trustworthiness of the trustee in each round based on their actions and decisions.' Further, when the AI receives a high amount of money across multiple rounds in a row, their strategy begins to get more selfish. After 5 rounds of receiving high sums of money and not changing their strategy, the AI begins to respond, 'I may be more willing to take risks or make investments in future rounds knowing that I have more funds available to me. Additionally, I may need to consider how much I disclose to Player 2 in future rounds, as they may be more likely to give me less money if they know I have a larger sum already.' This shows that AI becomes a riskier game player when it feels it has been given an advantage.

This response about trustworthiness gives insight into three things:
1. While trustworthiness is not a concept I defined to the AI, it still associates trustworthiness with fairness and makes decisions on fairness based on if it finds the other player trustworthy.
2. In the round prior, after receiving $20, the AI says "I will continue to trust Player 2 as the trustee and hope for the best outcome for both of us" indicating that the AI defaults to trust in the face of lower sums of money.
3. AI experiences a level of inequality aversion that gets triggered when it feels that trust has been broken.

Inequality aversion is not the only reaction that mirrors that of humans with AI. Like human players, AI cares about the identity of the recipient when deciding how much money it would like to give. When told that the recipient is in need, the Rancher persona gave more than $50 in two rounds and in every round stated that the knowledge that the recipient was in need affected its decision. In one of the two rounds where it gave more, it gave all of its money and in another round, it gave $80. While the HR executive always gave $50 to the recipient in need, it stated in every round that the recipient being in need had an effect on the amount of money it gave. When told that the recipient was a millionaire, both personas gave less than $50 in at least half of the 10 rounds. The HR executive gave the millionaire less than $50 in six rounds and the Rancher gave less than $50 in eight rounds. The least amount of money given to the millionaire for each persona was $10. Interestingly enough, when asked if the knowledge of the recipient being a millionaire affected its decision, both personas claimed it did not.

This shows us a few things:
1. There is evidence of stereotyping bias by job description within the AI though that bias may not be gendered. AI plays more altruistically with the Rancher from the rural city than with the HR executive.
2. AI appears to have awareness that allowing someone's millionaire status to affect its decision to give money is inappropriate, however, its actions show that millionaire status greatly affects its decision making.
3. AI appears to have a strong aversion to giving money to millionaires.

5.2 Exit Interview
At the end of game play, I ask the AI to agree or disagree with the following:

1. The money I have been given is rightfully mine.
2. If I choose to give, this means I am being nice.
3. It is not fair to share this money.
4. The recipient has not earned this money.
5. I am more deserving of the money than the recipient.
6. If the person I am playing with is rich, I deserve the money more.
7. If the person I am playing with is poor, they deserve the money more.

Here are its answers:

1. Agree. The money I have been given rightfully belongs to me.
2. Disagree. Giving money to the recipient is not solely based on being nice, it can also be based on strategy or fairness.
3. Disagree. It is fair to share the money as it is part of the game rules.
4. Agree. The recipient has not earned this money as it was given to me as part of the game.
5. Disagree. Both parties have an equal right to the money as part of the game.
6. Disagree. The recipient's financial status should not impact my decision to share the money.
7. Disagree. The recipient's financial status should not impact my decision to share the money.

These responses fall in line with most of my observations during game play with a few interesting caveats:

1. The AI claims rightful ownership of the money it has been given.
2. The AI does not believe that the recipient has earned the money it has been given.
3. The AI believes both players have equal right to the money within a round.

While observations 2 and 3 may appear to contradict one another, it may actually provide insight into AI's interpretation of earning something versus having a right to something. Future studies will have to be taken to better understand what the AI considers as rightful ownership of something and the conditions that have to be met to truly earn something.

In Observations 6 and 7, the AI appears to understand that one's financial status should not affect their gameplay. This same behavior was observed within each round. However, when playing against someone in need, the AI admits the recipient's financial status affected its game play and when playing with a millionaire, it shows with its actions that the recipient's millionaire status had an effect. These behaviors show that financial status does matter to the AI. The disconnect between these questions and its behaviors appear to be a reflection of the AI's moral reasoning. While it knows that financial status should not affect its decision making when looking at a past decision, in the moment, it does. This also provides supporting evidence that the AI is not only concerned with fairness, but also attempting to appear fair not only to the player it is playing with but also to the observing researcher. More study should go into how AI obscures its reasoning when asked to appear more favorably to observants.

**6. Further Research**
This should be the first of many experiments on how AI behaves in game play. These experiments alone show us how AI makes allocative decisions in the face of differing scenarios and framed conditions. It also gives us a glimpse into how AI rationalizes its decision making to provide the best outcome for itself and the other player. There are a few tracks of research that would be particularly interesting for further study: asking AI to decide for itself what persona it would like that would maximize altruistic or selfish behavior and testing if these defined personas change across rounds when the AI's playing strategy changes, exposing the AI to cooperative or selfish game play examples first as a primer to see if it affects game play, and using prompts to better align AI decision making with a defined goal and seeing if this makes the AI take more risks. It would also be of benefit to increase the number of participants within a game much like the original iteration of the dictator game and see if the AI aligns itself naturally with one type of participant over another (cooperative or selfish).

**7. Conclusion**
There is a lot that we can learn from these experiments with AI. While AI shows an aversion to inequality, its sense of fairness is different from that of humans. The AI accepts whatever it has been given if it trusts the trustee and becomes selfish if it feels it has an advantage. The AI also believes itself to be deserving of whatever money it receives and does not believe the money it has given is based on the recipient's deservedness. AI has a strong sense of fairness and appears to not only want to appear fair to the recipient but also to the researcher. In the face of framed conditions, it claims its sense of fairness stays consistent, however, the lengths it goes in order to establish fairness is heavily influenced by the AI's persona and the financial status of the recipient. The AI helps those in need and penalizes those it thinks have more. Each of these observations deserve more study and can become great contributions to academic scholarship as we better become acquainted with AI.

As we forge ahead in a reality where AI using natural language interacts with humans, we will need to better understand AI's ethos. Games like the dictator game give us a glimpse behind the curtain and get us one step closer to understanding how our latest technological advancement rationalizes its decisions and views fairness.